\documentclass[10pt,twocolumn,letterpaper]{article}

\usepackage{amsmath,amsfonts,bm}

\def\eqref#1{equation~\ref{#1}}

\def\1{\bm{1}}

\DeclareMathAlphabet{\mathsfit}{\encodingdefault}{\sfdefault}{m}{sl}
\SetMathAlphabet{\mathsfit}{bold}{\encodingdefault}{\sfdefault}{bx}{n}

\def\eg{\textit{e.g.,~}}

\def\etal{\textit{et al.~}}

\usepackage{cvpr}              

\usepackage[accsupp]{axessibility}
\usepackage{graphicx}
\usepackage{amsmath}
\usepackage{amssymb}
\usepackage{booktabs}
\usepackage{multirow}
\usepackage{makecell}
\usepackage{colortbl}
\usepackage[dvipsnames]{xcolor}

\usepackage[dvipsnames]{xcolor}
\newcommand\modified[1]{\textcolor{black}{#1}}
\newcommand\final[1]{\textcolor{black}{#1}}

\definecolor{myGreen}{RGB}{72,180,75}
\definecolor{myBlue}{RGB}{78,123,232}
\definecolor{myRed}{RGB}{236,76,76}

\newcommand{\wwj}{\textcolor{black}}

\newcommand{\lrd}{\textcolor{black}}

\usepackage[pagebackref,breaklinks,colorlinks]{hyperref}

\usepackage[capitalize]{cleveref}
\crefname{section}{Section}{Secs.}
\Crefname{section}{Section}{Sections}
\Crefname{table}{Table}{Tables}
\crefname{table}{Table}{Tabs.}

\def\cvprPaperID{5714} 
\def\confName{ICCV}
\def\confYear{2023}

\begin{document}

\title{Similarity Min-Max: Zero-Shot Day-Night Domain Adaptation}

\author{Rundong Luo$^{1,2,3}$ \hspace{1mm} Wenjing Wang$^{1}$ \hspace{1mm} Wenhan Yang$^{4}$ \hspace{1mm} Jiaying Liu$^{1}$\thanks{Corresponding author.} \\
$^{1}$ Wangxuan Institute of Computer Technology, Peking University \\
$^{2}$School of EECS, Peking University \hspace{1mm} $^{3}$School of CS, Peking University
 \hspace{1mm} $^{4}$ Peng Cheng Laboratory
}
\maketitle


\begin{abstract}
Low-light conditions not only hamper human visual experience but also degrade the model's performance on downstream vision tasks.
While existing works make remarkable progress on day-night domain adaptation, they rely heavily on domain knowledge derived from the task-specific nighttime dataset.
This paper challenges a more complicated scenario with border applicability, i.e., \textbf{zero-shot} day-night domain adaptation, which eliminates reliance on any nighttime data.
Unlike prior zero-shot adaptation approaches emphasizing either image-level translation or model-level adaptation, we propose a similarity min-max paradigm that considers them under a unified framework.
On the image level, we darken images towards minimum feature similarity to enlarge the domain gap.
Then on the model level, we maximize the feature similarity between the darkened images and their normal-light counterparts for better model adaptation.
To the best of our knowledge, this work represents the pioneering effort in jointly optimizing both levels, resulting in a significant improvement of model generalizability.
Extensive experiments demonstrate our method's effectiveness and broad applicability on various nighttime vision tasks, including classification, semantic segmentation, visual place recognition, and video action recognition. 
Our project page is available at \url{https://red-fairy.github.io/ZeroShotDayNightDA-Webpage/}.
\end{abstract}

\section{Introduction}
\label{sec:intro}
Deep neural networks are sensitive to insufficient illumination, and such deficiency has posed significant threats to safety-critical computer vision applications. Intuitively, insufficient illumination can be handled by low-light enhancement methods~\cite{URetinexNet, enlightenGAN, Zero-DCE++, SCI, LEDNet, D2HNet}, which aim at restoring low-light images to normal-light. However, enhancement models do not necessarily benefit downstream high-level vision tasks as they are optimized for human visual perception and neglect the need for machine vision.

\begin{figure}
    \centering
    \includegraphics[width=\linewidth]{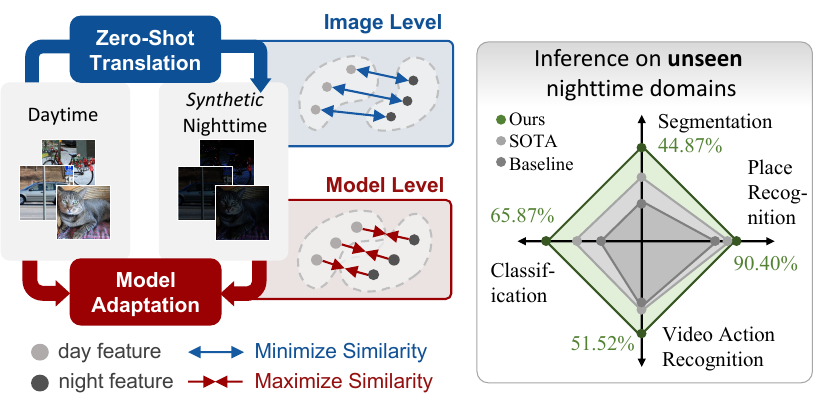}
    \caption{Left: Illustration of our similarity min-max framework for zero-shot day-night domain adaptation. Right: Our framework achieves state-of-the-art results on multiple downstream high-level vision tasks without seeing real nighttime images during training.
    }
    \label{fig:teaser}
\end{figure}

Much existing literature has focused on improving machine vision performance at night through domain adaptation. By aligning the distribution statistics between the nighttime and daytime datasets through image translation~\cite{car-detection, bridge-day-night-gap-segmentation,cross-domain-distill}, self-supervised learning~\cite{HLA-face-v2, SACC}, or multi-stage algorithms~\cite{GCMA, yolo-in-the-dark,nightlab}, these methods have greatly improved models' performance in nighttime environments.
The primary assumption of domain adaptation is that the target domain data is readily available.
\final{Nevertheless, obtaining data from the task-specific target domain may be challenging in extreme practical application scenerios such as deep-space exploration and deep-sea analysis.}

To reduce the requirement on target domain data, \textbf{zero-shot} domain adaptation has emerged as a promising research direction, where adaptation is performed without accessing the target domain.
Regarding day-night domain adaptation, the primary challenge is learning illumination-robust representations generalizable to both day and night modalities.
To accomplish this goal under zero-shot constraints, Lengyel~\etal~\cite{CIConv} proposed a color invariant convolution for handling illumination changes.
Cui~\etal~\cite{MAET} designed a Reverse ISP pipeline and generated synthetic nighttime images with pseudo labels.
\final{However, image-level methods simply consider synthetic nighttime as pseudo-labeled data and overlook model-level feature extraction; model-level methods focus on adjusting model architecture but neglect image-level nighttime characteristics.
Neither is effective enough 
capture the illumination-robust representations that could bridge the complex day-night domain gap.}

From this point of view, we devise a similarity min-max framework that involves two levels, as illustrated in Figure~\ref{fig:teaser}.
On the image level, we generate a synthetic nighttime domain that shares minimum feature similarity with the daytime domain to enlarge the domain gap.
On the model level, we learn illumination-robust representations by maximizing the feature similarity of images from the two domains for better model adaptation.

Intuitive as it seems, solving this bi-level optimization problem is \textbf{untrivial}.
Directly solving it may yield unsatisfactory results, \eg, meaningless images filled with zero values or identical features given all inputs. 
\final{
Therefore, we develop a stable training pipeline that can be considered a sequential operation on both the image and the model. Regarding the image, we propose an exposure-guided module to perform reliable and controllable nighttime image synthesis. Regarding the model, we align the representation of images from day and night domains through multi-task contrastive learning.
Finally, our model achieves day-night adaptation without seeing real nighttime images.}

Our framework can serve as a plug-and-play remedy to existing daytime models. To verify its effectiveness, we conduct extensive experiments on multiple high-level nighttime vision tasks,
including classification, semantic segmentation, visual place recognition, and video action recognition. Results on various benchmarks demonstrate our superiority over the state-of-the-art.

Our contributions are summarized as follows:
\begin{itemize}
    \item We propose a similarity min-max framework for zero-shot day-night domain adaptation. Feature similarity between the original and darkened images is minimized by image-level translation and maximized by model-level adaptation. In this way, model's performance in nighttime is improved without accessing real nighttime images.
    \item We develop a stable training pipeline to solve this bi-level optimization problem. On the image level, we propose an exposure-guided module to perform reliable and controllable nighttime image synthesis. On the model level, we align the representation of images from day and night domains through multi-task contrastive learning.
    \item Our framework universally applies to various nighttime high-level vision tasks.
    Experiments on classification, semantic segmentation, visual place recognition, and video action recognition demonstrate the superiority of our method.
\end{itemize}

\section{Related Works}
\label{sec:related-works}

\noindent \textbf{Low-Light Enhancement.}
A straightforward approach to improve the model's performance in low light is brightening the test low-light images. Early non-learning practices exploit image processing tools such as histogram equalization~\cite{histrogram-equalization} or image formation theories such as Retinex Theory~\cite{Retinex-original}. Recent literature mainly takes advantage of the advance in deep learning. Trained on paired day-night data, some methods~\cite{RetinexNet,URetinexNet,RUAS} simulate the image decomposition process of Retinex Theory.
Others introduce adversarial learning~\cite{enlightenGAN} to support unpaired training. Zero-DCE ~\cite{Zero-DCE,Zero-DCE++} designs a curve-based low-light enhancement model and trains in a zero-reference way. Advanced techniques, including frequency decomposition~\cite{LightSuppression}, feature pyramids~\cite{D2HNet,LEDNet}, and flow models~\cite{llflow} are also adopted in recent papers.

\noindent\textbf{Day-Night Domain Adaptation.}
Nighttime high-level vision has attracted increasing attention in recent years. Apart from pre-processing with enhancement models, day-night domain adaptation is also a viable solution. YOLO-in-the-dark~\cite{yolo-in-the-dark} introduces the glue layer to mitigate the day-night domain gap. 
MAET~\cite{MAET} exploits image signal processing (ISP) for nighttime image generation and uses both synthetic and real nighttime images for training.
HLA-face~\cite{HLA-face-v2} proposes a joint high-low adaptation framework driven by self-supervised learning. 
Others~\cite{GCMA, DANNET, car-detection, bridge-day-night-gap-segmentation, GAN-DA4driving} employ Generative Adversarial Network (GAN) to transfer labeled daytime data to nighttime.

\noindent\textbf{Zero-Shot Day-Night Domain Adaptation.} 
\modified{Beyond Conventional adaptation, \textbf{zero-shot} approaches consider an even stricter condition where real nighttime images are inaccessible. For general tasks, existing methods either draw supports from extra task-irrelevant source and target domain data pairs~\cite{zero-shot-1,zero-shot-2} or require underlying probability distribution of the target domain~\cite{zero-shot-3}, which are inapplicable to our settings.
For the day-night task, Lengyel \textit{et al.} propose the Color Invariant Convolution (CIConv)~\cite{CIConv} to capture illumination-robust features. MAET \cite{MAET} can be viewed as zero-shot when real nighttime images are discarded during finetuning. Besides, domain generalization methods~\cite{AdaBN,IRM,MixStyle,RobustNet,SAN-SAW,FSDR,WEDGE} also apply to our settings since they do not know target domains, but they are too general to handle the complex day-night domain gap.
}

Despite these advances, low-light enhancement concentrates on human vision and disregards downstream nighttime vision tasks. Conventional adaptation methods require task-specific nighttime datasets, which creates extra burdens on data collection and limits their generalizability to multiple tasks. Prior zero-shot adaptation methods fail to consider image-level and model-level jointly. In this paper, we propose a novel similarity min-max framework that could outperform existing methods by a large margin.



\section{Similarity Min-Max Optimization}

This section introduces our approach for zero-shot day-night domain adaptation. We first explain our motivation and then introduce the overall framework and detailed designs.

\subsection{Motivation}
\label{sec:motivation}
\begin{figure}
    \centering
    \begin{minipage}{\linewidth}
        \includegraphics[width=\linewidth]{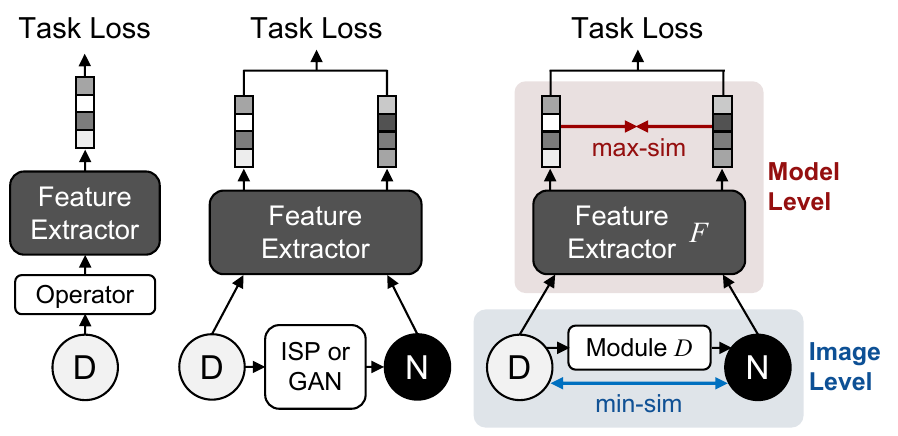}
    \end{minipage}
    \begin{minipage}{\linewidth}
        \footnotesize \hspace{18pt}(a)\hspace{52pt}(b) \hspace{75pt}(c) 
    \end{minipage}
    \caption{Comparison between different learning paradigms.
    \textsf{D} and \textsf{N} denote the daytime and nighttime domains, respectively. (a) Operator-based. (b) Darkening-based. (c) Our method.}
    \label{fig:comparison}
\end{figure}

\wwj{Existing methods, generally categorized into Operator-based and Darkening-based as shown in Figure~\ref{fig:comparison}, come across troubles in the day-night domain adaptation problem.}
Operator-based methods~\cite{CIConv} rely on the manually designed operators \textit{at the model level} to handle illumination variations, which are not adaptive to real complex scenarios.
Darkening-based methods transfer labeled daytime data to nighttime by ISP~\cite{MAET} or GAN~\cite{I2I-darken, GCMA, car-detection, bridge-day-night-gap-segmentation} only \textit{at the image level}.
However, the former is sensor-dependent and cannot generalize across devices and datasets, while the latter requires data from the task-specific nighttime domain and thus fails to generalize to our zero-shot setting. 

Intrinsically, the most critical issue of existing methods is their ignorance of the mutual effect between \textbf{pixels} and \textbf{features}.
In our work, we make the first systematic investigation on this issue and propose a similarity min-max framework that thoroughly exploits the information from two sides.
In detail, \wwj{\textit{at the pixel} (\textit{image}) \textit{level}}, we minimize the feature similarity between original and darkened images by day-to-night translation. While \wwj{\textit{at the feature} (\textit{model}) \textit{level}}, we maximize the feature similarity by representation alignment. 
This joint optimization leads to representations more robust to illumination changes.

We formulate our framework as follows. Denote the feature extractor of the downstream model as $F(\cdot)$. Being robust to illumination requires the extracted feature of a daytime image $I$ and its nighttime version $D(I)$ to be similar, where $D(\cdot)$ represents a darkening process.
The limitation of existing darkening-based methods is that their $D$ does not consider the co-effect of $F$.
So we introduce additional constraints on $D$: we require $D$ to minimize the similarity between the day feature $F(I)$ and the night feature $F(D(I))$.
This way, we guide the darkening process with high-level vision, forming a unified framework of $D$ and $F$.
At this point, we can integrate $D$ and $F$ as a min-max optimization problem:
\begin{equation}
    \mathop{\max}_{\theta_F}\mathop{\min}_{\theta_D} \quad \operatorname{Sim}(F(I),F(D(I))),
    \label{eq:basic}
\end{equation}
where $\theta_D$ and $\theta_F$ denote the parameters in $D$ and $F$, and $\operatorname{Sim}(\cdot,\cdot)$ measures the similarity between features.

However, trivial solutions exist in Eq.~(\ref{eq:basic}), such as $D$ generating entirely black images and $F$ extracting identical features for all inputs. We add regularizations to $D$ and $F$ accordingly to address this problem:
\begin{equation}
    \mathop{\max}_{\theta_F}\mathop{\min}_{\theta_D} \ \operatorname{Sim}(F(I),F(D(I))) + \mathcal{R}_D(\theta_D) - \mathcal{R}_F(\theta_F) ,
    \label{eq:min-max}
\end{equation}
where $\mathcal{R}_D$ and $\mathcal{R}_F$ are intended to prevent model collapse.

How to design $\mathcal{R}_D$ and $\mathcal{R}_F$ properly is the key to solving Eq.~(\ref{eq:min-max}). The following will introduce how we design $\mathcal{R}_D$ and $\mathcal{R}_F$ and build up the whole learning framework.

\label{sec:method}

\begin{figure*}
  \centering
  \includegraphics[width=\linewidth]{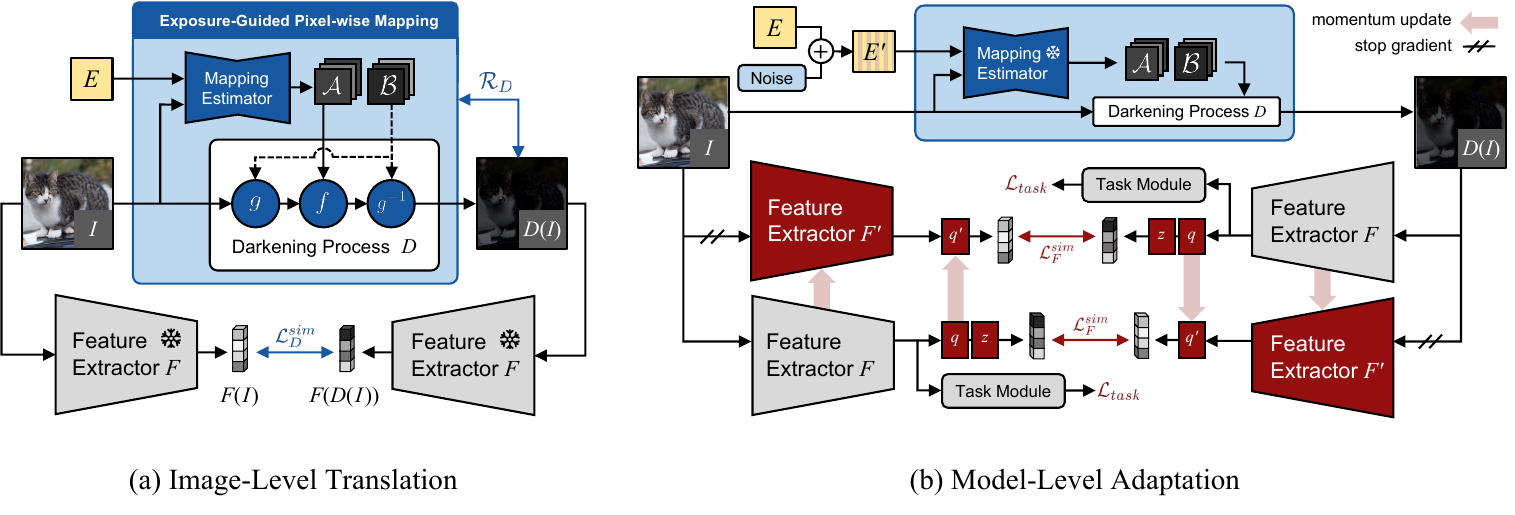}
  \caption{
  Our proposed similarity min-max framework for zero-shot day-night domain adaptation.
  (a) We first train a darkening module $D$ with a fixed feature extractor to generate \textit{synthesized} nighttime images that share minimum similarity with their daytime counterparts.
  (b) After obtaining $D$, we freeze its weights and maximize the day-night feature similarity to adapt the model to nighttime.
  }
  \label{fig:pipeline}
\end{figure*}

\subsection{Image-Level Similarity Minimization}
\label{sec:image-level}

This section describes our design for the darkening module $D$. We want $D$ to satisfy three properties:
\begin{itemize}
    \item \textbf{Stability}. First and foremost, we need to prevent the similarity min-max optimization from collapsing, \ie, applying proper $\mathcal{R}_D$ in Eq.~(\ref{eq:min-max}).
    \item \textbf{Generalization}. $D$ should represent a generalized darkening process so the downstream model can learn useful knowledge from $D(I)$ to handle unseen nighttime scenes.
    \item \textbf{Flexibility}. \final{We additionally expect flexible control over the degree of darkening, which could enable us to create diverse inputs beneficial for optimizing $F$.}
\end{itemize}

We design an \modified{exposure-guided pixel-wise mapping algorithm to satisfy the above properties.
Unlike widely-used image-to-image darkening approaches~\cite{I2I-darken,GCMA,car-detection} that rely heavily on real nighttime images, pixel-wise mapping adjusts images using a pre-selected function with learnable parameters.
}
%
%
We empirically found that, by setting proper constraints on \modified{the mapping function}, we can naturally avoid obtaining trivial solutions in the similarity min-max optimization (\textit{stability}) and guarantee $D$ follows a typical low-light process (\textit{generalization}).
Finally, we add an exposure guidance mechanism for better \final{\textit{flexibility}}. The detailed design will be illustrated as follows.

\vspace{1mm}
\noindent \textbf{Darkening Process.} 
\wwj{We first define a general function for tone mapping.
Given an image $I \in [0,1]^{C \cdot H \cdot W}$, we use a non-linear mapping \wwj{$f$: $[0,1]\rightarrow[0,1]$} and a pixel-wise adjustment map $\mathcal{A} \in [0,1]^{C \cdot H \cdot W}$ to process the image:}
\begin{equation}
    D^{0}(I)=f(I,\mathcal{A}).
\end{equation}
\wwj{Typically, $f$ should be monotonically increasing to preserve contrast and satisfy $f(1,\alpha)=1$ for all $\alpha$ to avoid information loss (\eg, gamma correction).}
However, the latter constraint $f(1,\alpha)=1$ no longer holds for darkening. Therefore, we propose an auxiliary pixel-wise adjustment using a monotonic increasing function $g$: $[0,1]\rightarrow[0,1]$ parameterized by another adjustment map $\mathcal{B} \in [0,1]^{C \cdot H \cdot W}$. \lrd{Note that $g$ only serves as a complement and should be simple to avoid taking over the role of $f$.} The overall darkening process is formulated as:
\begin{align}
    &D(I) = g^{-1}(f(g(I,\mathcal{B}),\mathcal{A}),\mathcal{B}).
\label{eq:darkening}
\end{align}
Both $\mathcal{A}$ and $\mathcal{B}$ are estimated by a mapping estimator conditioned on the input image $I$.

\lrd{
To guarantee $D$ represents a darkening process (\ie, $D(I) < I$), $f$ should additionally satisfy convexity. Specifically, we let $f$ be the iterative quadratic curve~\cite{Zero-DCE}: $f(x)$~$=$~$h^{(8)}(x)$, $h(x,\alpha)$ $=$~$\alpha x^2 + (1-\alpha)x$, and $g$ be the dividing operation: $g(x, \beta)$~$=$~$x/\beta$ in our implementation.
}
Other kinds of curve forms are also considered and tested.
Still, we empirically found that quadratic curves could bring slightly better results (results in Sec.~\ref{sec:image-classification}).
%

Besides, to enable flexible control over the exposure level, we feed an exposure map $E$ to the mapping estimator with $I$, yielding the corresponding darkened image $D(I,E)$. During training, the darkening module is encouraged to align the pixel value of $E$ and $D(I,E)$. We use $D(I)$ and $D(I,E)$ interchangeably for simplicity.

\vspace{1mm}
\noindent \textbf{Similarity Minimization.}
The training objective of module $D$ involves two parts:
similarity minimization and regularization.
For the former, we directly reduce the distance between features:
\begin{align}
    \mathcal{L}^{sim}_D = \frac{\langle F(I), F(D(I)) \rangle}{||F(I)||_2\cdot ||F(D(I))||_2},
    \label{eq:sim-D}
\end{align}
where $\langle\cdot,\cdot\rangle$ is the inner product between two vectors.

The regularization term consists of four losses. Besides a color consistency loss $\mathcal{L}_{col}$~\cite{Zero-DCE} that corrects color deviations, three additional losses are proposed to regularize $D$:

Firstly, conditional exposure control is adopted to align the exposure map with the corresponding generated image:
\begin{equation}
    \mathcal{L}_{c-exp} = \sum_{1\leq i\leq H,\\1\leq j\leq W}|\hat{D}_{i,j}(I,E) - E_{i,j}|,
    \label{eq:exp}
\end{equation}
where $\hat{D}(I,E)$ is the channel-wise average of $D(I,E)$. During training, each exposure map $E$ has identical entries uniformly sampled between $[0,0.5]$.

\modified{Then we add constraints on $\mathcal{A}$.} \wwj{Intuitively, $\mathcal{A}$ represents the degree of illumination reduction. Illumination usually varies slowly across a scene but encounters rapid variations from object to object. Following this property, we apply a loose total variance loss:}
\begin{align}
    \mathcal{L}_{ltv}(\mathcal{A}) & =  \sum_{c\in\{R,G,B\}}(h(|\nabla_x\mathcal{A}^c|)^2 + h(|\nabla_y\mathcal{A}^c|)^2), \, 
    \label{eq:tv} \\
    h(x) & = \max(\alpha - |x-\alpha|, 0),
\end{align}
where $\nabla_x, \nabla_y$ are gradient operations along the horizontal and vertical axis, respectively, and $\alpha$ is a hyperparameter.
Compared with the original total variance loss where $h$ is the identity function, our loose version allows the network to predict values of greater difference for adjacent pixels, which is common on objects' boundaries.

Finally, we adopt $\mathcal{L}_{flex}(\mathcal{B}) = 1-\mathcal{B}$ to avoid model fitting to the exposure solely by $g$.

The overall training objective for $D$ is:
\begin{align}
    \mathcal{L}_{D} &= \lambda^{sim}_D\mathcal{L}^{sim}_D + \mathcal{R}_D, \\
    \mathcal{R}_D  &= \lambda_{c-exp}\mathcal{L}_{c-exp} + \lambda_{col}\mathcal{L}_{col} + \lambda_{ltv}\mathcal{L}_{ltv} +
    \lambda_{flex}\mathcal{L}_{flex}.
    \label{eq:loss-D}
\end{align}

\subsection{Model-Level Similarity Maximization}
\label{sec:model-level}
The darkening module $D$ grants us access to a synthetic nighttime domain. In this section, we exploit $D$ to learn illumination-robust representations.

Contrastive learning~\cite{simclr,moco} is a self-supervised learning paradigm that contrasts positive and negative image pairs. However, images of the same class in classification or adjacent scenes in segmentation will form false negative pairs, thus hurting the model's performance. 
To alleviate these burdens, BYOL~\cite{BYOL} proposes a non-negative variant that only aligns the feature between positive image pairs $\{v,v^+\}$:
\begin{equation}
    \mathcal{L}_{\text{BYOL}}(v,v^+) = 2 - \frac{2 \cdot \langle z(q(F(v))), q'(F'(v^+)) \rangle}{||z(q(F(v)))||_2 \cdot ||q'(F'(v^+))||_2},
    \label{eq:BYOL}
\end{equation}
where $q, q'$ are projection heads, and $z$ is the prediction head. Both of them are MLPs with a single hidden layer.
Note that $F'$ and $q'$ share the same architecture and weight initialization with $F$ and $q$ but receive no gradient and are updated by exponential moving average (EMA).

\vspace{1mm}
\noindent \textbf{Similarity Maximization.}
Motivated by BYOL, we maximize the feature similarity between synthetic nighttime and daytime domains by non-negative contrastive learning.
Given a daytime image $I$ and an exposure map $E$, we formulate the training objective as follows:
\begin{equation}
    \mathcal{L}^{sim}_F = \mathcal{L}_{\text{BYOL}}(I, D(I,E)) + \mathcal{L}_{\text{BYOL}}(D(I,E),I).
    \label{eq:sim-F}
\end{equation}
Note that the measure of feature similarity is different between Eq.~(\ref{eq:sim-D}) and Eq.~(\ref{eq:sim-F}). Directly applying Eq.~(\ref{eq:sim-D}) to train $F$ brings poorer results due to potential feature degeneration.
In comparison, the asymmetric projection head and stop gradient policies prevent the feature extractor $F$ from collapsing, \ie, working as the regularization $\mathcal{R}_F$ in Eq.~(\ref{eq:min-max}) together with the task loss (introduced below).

Moreover, different from $E$ in Eq.~(\ref{eq:exp}), we use a compound exposure map $E'$ instead.
$E'$ is first initialized with identical entries uniformly sampled between $[0,0.2]$ for simulating nighttime illumination.
This range is the same for all downstream tasks, which does not introduce task-relevant prior.
Then, we add pixel-wise noise $z_1$ and patch-wise noise $z_2$ to $E$ to simulate exposure discrepancy. Overall, $E'$ can be represented as:
\begin{equation}
    E' = \mathcal{U}(0, 0.2) + z_1 + z_2.
\end{equation}
See the supplementary for details on noise injection.

Besides $\mathcal{L}^{sim}_F$, we add task-specific supervision $\mathcal{L}_{task}$ on both the original daytime and synthetic nighttime domain.
The final training objective for $F$ is:
\begin{align}
    \mathcal{L}_{F} &= \lambda^{sim}_F\mathcal{L}^{sim}_F + \lambda_{task}\mathcal{L}_{task}.
    \label{eq:loss-F}
\end{align}

\begin{figure}
    \centering
    \includegraphics[width=\linewidth]{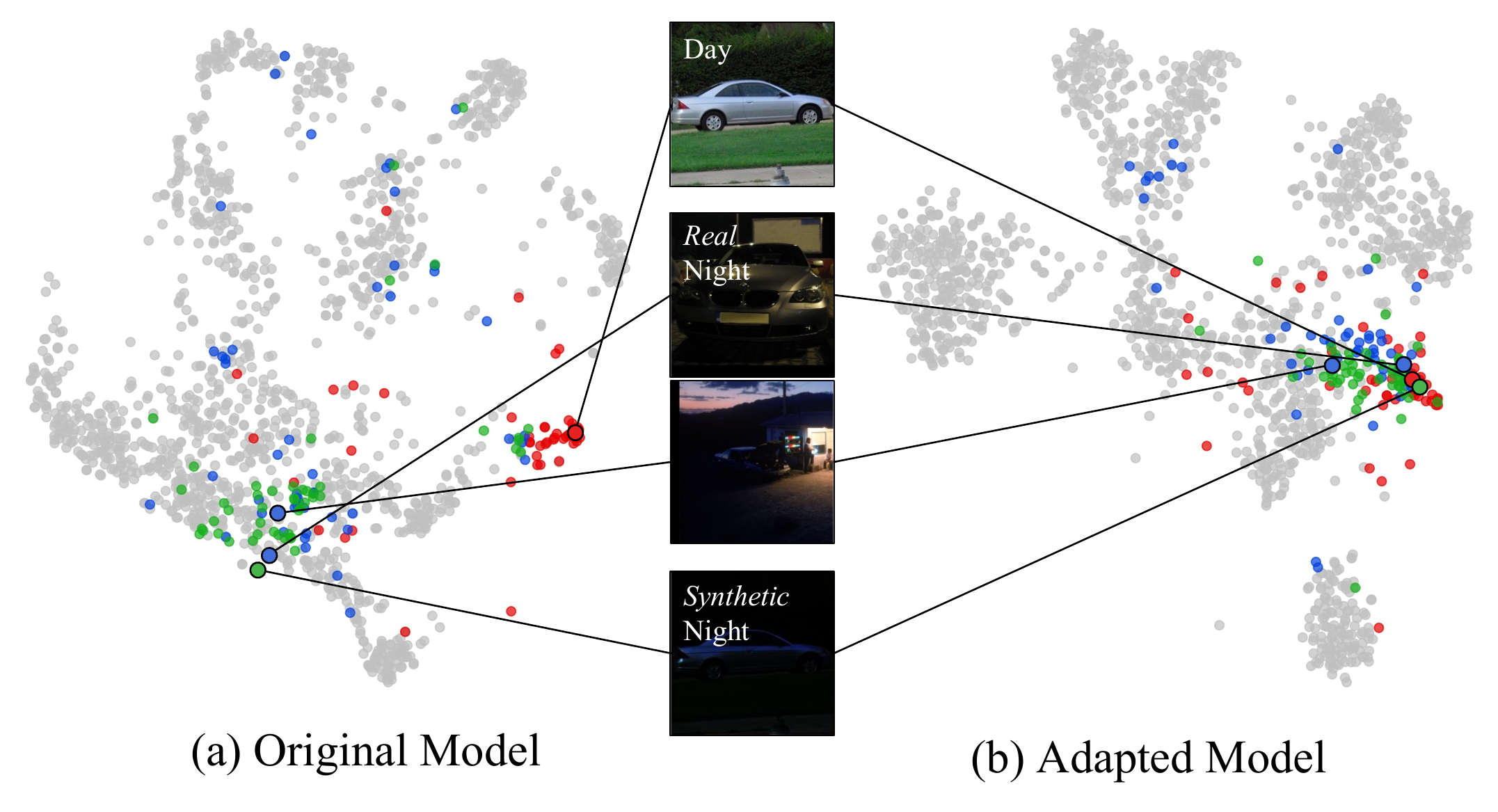}
    \caption{t-SNE~\cite{tsne} visualization of images' feature extracted by the original daytime model and our adapted model on CODaN~\cite{CIConv}. {\textcolor{myRed}{Red}}, {\textcolor{myGreen}{green}}, and {\textcolor{myBlue}{blue}} dots represent the feature of daytime, \textit{synthesized} nighttime, and \textit{real} nighttime images, respectively. We only color the instances from the ``Car'' category for better visual quality. Additional visualization results are shown in the supplementary.
    }
    \label{fig:tsne-visual}
\end{figure}

\subsection{Overall Training Pipeline}

Having introduced the image-level similarity minimization (Section \ref{sec:image-level}) and model-level similarity maximization (Section \ref{sec:model-level}), this section discusses the overall pipeline, as shown in Figure~\ref{fig:pipeline}.

An intuitive idea is training $D$ and $F$ alternately like GAN~\cite{GAN,CycleGAN}.
Nevertheless, balancing $D$ and $F$ increases the difficulty of parameter tuning and makes the optimization process unstable.
We adopt a simple but effective two-step strategy to solve this problem: we first train $D$ and keep $F$ frozen, then train $F$ and keep $D$ frozen.
Compared with the alternate strategy, our step-wise approach improves the performance on nighttime image classification (elaborate in Section \ref{sec:image-classification}) from 63.84\% to 65.87\%.

We could also explain the merits of our min-max framework from the perspective of adversarial training~\cite{FGSM,DYNACL}. Module $D$ first produces the worst-case examples regarding feature similarity.
Then, our model could learn the illumination-robust features by training on these cases through similarity maximization. This technical commonality further justifies our motivation to build the similarity min-max framework.

Across all downstream tasks, the feature extractor and task module are initialized by daytime pre-trained models. We first freeze the feature extractor and train the darkening module (image-level translation). Then, we keep the darkening module fixed and train the feature extractor and task module jointly (model-level adaptation).

\subsection{Empirical Justifications on Darkening Module}
\final{Simulating nighttime conditions without accessing real nighttime images is the key to our framework.} Particularly, nighttime conditions bring semantic changes in addition to illumination changes, \eg, the dark environment with artificial lights on the second real nighttime image in Figure~\ref{fig:tsne-visual}.
However, an accurate simulation is extremely difficult since our prior knowledge is limited to ``low illumination''.
Fortunately, unlike typical day-to-night image synthesis processes~\cite{neural-ISP} which target the human visual experience, ours only care about the distribution of darkened images in the feature space.
Leaving aside visual quality, we are pleased to find that the feature distribution of our synthesized nighttime domain is similar to that of the real nighttime domain as visualized in Figure~\ref{fig:tsne-visual}(a).
This observation demonstrates that our darkening process can characterize the night domain from the model-level perspective.

Thanks to this property, the feature discrepancy between daytime and \textit{real} nighttime domain is significantly reduced
after model-level adaptation (red and blue dots in Figure~\ref{fig:tsne-visual}).
This discovery is consistent with the Maximum Mean Discrepancy (MMD) between the feature distribution of day and night modalities, which is 0.020 and 0.014 for the original and adapted models, respectively. We provide implementation details and additional empirical analysis using saliency maps in the supplementary.


\section{Experiments}
\label{sec:experiments}
This section provides the implementation details, benchmarking results, and ablation analysis of our method.

\subsection{Implementation Details}
\label{sec:implementation-details}
Our framework widely applies to various nighttime vision tasks. In the following, we evaluate our method with four representative tasks: image classification, semantic segmentation, visual place recognition, and video action recognition.
Only daytime data are accessible for training and validation, while nighttime data are only used during evaluation.
\lrd{
We benchmark our method with three categories of methods that require no dataset-specific target domain data: low-light enhancement, zero-shot day-night domain adaptation, and domain generalization.} 
For low-light enhancement, enhancement models are trained on their original datasets. Then we adopt them as a pre-processing step to assist the daytime baseline. The results of our method is the average of three independent trails. Additional details are provided in the supplementary. 

\subsection{Nighttime Image Classification}
\label{sec:image-classification}

We first consider one of the most fundamental vision tasks: image classification. CODaN~\cite{CIConv} is a 10-class dataset containing a training set of 10000 daytime images and a test set with 2500 daytime and nighttime images, respectively. We validate models on the daytime test set and evaluate them on the nighttime test set. The backbone is ResNet-18~\cite{ResNet}.

Benchmarking results are shown in Table~\ref{tab:classification}. Enhancement methods restore input low-light images from the human visual perspective while keeping the model untouched, resulting in limited performance gains.
Domain generalization methods are designed for general tasks and perform poorly in unseen nighttime environments.
%
MAET~\cite{MAET} relies on degrading transformation with sensor-specific parameters, which suffers from poor generalizability.
CIConv~\cite{CIConv} adopts learnable color invariant edge detectors, which are not robust to the complex illumination variation in real scenarios.
In contrast, our method outperforms state-of-the-art methods by a large margin (60.32\%~v.s.~65.87\%), demonstrating our unified framework could obtain features more robust to illumination shifts.
%

\begin{table}[t]
    \centering
    \small
    \caption{
    Top-1 classification accuracy on the CODaN nighttime test set~\cite{CIConv}. $\dagger$ denotes our re-implementation with both the original and synthesized image fed into the task module.}
    \begin{tabular}{l|c}
    \toprule
    Method  & \makecell[c]{Top-1 (\%)}\\ \midrule
    ResNet-18~\cite{ResNet}  & 53.32\\ \midrule
    \multicolumn{1}{l}{\textbf{Low-Light Enhancement}} & \\ \midrule
    EnlightenGAN~\cite{enlightenGAN} & 56.68\\
    LEDNet~\cite{LEDNet} & 57.40 \\
    Zero-DCE++~\cite{Zero-DCE++} & 57.96\\ 
    RUAS~\cite{RUAS} & 58.36 \\
    SCI~\cite{SCI} & 58.68\\ 
    URetinexNet~\cite{URetinexNet} & 58.72\\\midrule
    \multicolumn{1}{l}{\textbf{Domain Generalization}} & \\ \midrule
    MixStyle~\cite{MixStyle} &53.12\\
    IRM~\cite{IRM} & 54.52 \\
    AdaBN~\cite{AdaBN} & 54.25\\ \midrule
    \multicolumn{1}{l}{\textbf{Zero-Shot Day-Night Domain Adaptation}} & \\\midrule
    MAET$\dagger$~\cite{MAET} & 56.48\\
    CIConv~\cite{CIConv} & 60.32\\
    \textbf{Ours} & \textbf{65.87}\\
    \bottomrule
    \end{tabular}
    \label{tab:classification}
\end{table}

\begin{table}[t]
    \centering
    \small
    \caption{
    Ablation studies for module $D$ and similarity losses. We report the Top-1 accuracy on the CODaN~\cite{CIConv} nighttime test set.}
    \begin{tabular}{l|l|c}
    \toprule
    Category & Method & Top-1 (\%)\\
    \midrule
    Baseline & Vanilla ResNet-18 & 53.32 \\
    \midrule
    Module $D$ & Brightness adjustment & 57.96 \\
    {Heuristic}& Gamma correction & 63.96\\
    \midrule
    Module $D$ & Reciprocal curve & 62.60 \\
    {Learnable}& Gamma curve & 64.16\\
    \midrule
    \multirow{2}{*}{Similarity}& w/o $\mathcal{L}_D^{sim}$ and  $\mathcal{L}_F^{sim}$& 64.08 \\ 
    \multirow{2}{*}{Loss} & w/o $\mathcal{L}_D^{sim}$ & 64.56\\ 
    & w/o $\mathcal{L}_F^{sim}$ & 64.88 \\ 
    \midrule
    Full version & - & \textbf{65.87}\\
    \bottomrule
    \end{tabular}
    \label{tab:classification-ablation}
\end{table}

\noindent \textbf{Ablation Studies.} We conduct ablation studies to justify our framework design in Table \ref{tab:classification-ablation}.
Firstly, we study how to design the darkening module $D$ given $\mathcal{L}_D^{sim}$. The model-level adaptation stage (Section~\ref{sec:model-level}) remains the same for fair comparisons.
%
%
Firstly, we replace our darkening module with heuristic image adjustment approaches, such as brightness adjustment (\texttt{Brightness} in \texttt{PIL}\footnote{https://pillow.readthedocs.io/en/stable/reference/ImageEnhance.html}) and gamma correction ($D(I) = I^\gamma$).
We implement these two approaches using a fixed darkening hyperparameter chosen after multiple trials and report the best score.

\modified{Next, we test other possible curve forms for $f$. Both gamma curve ($f(x,\alpha)=x^\frac{1}{\alpha}, \alpha \in (0,1]$) and reciprocal curve ($f(x,\alpha) = \frac{(1-\alpha)\cdot x}{1-\alpha\cdot x}, \alpha \in [0,1)$) bring slightly worse results than the iterative quadratic curve.
Please refer to the supplementary for implementation details of these ablations and additional results on the segmentation task.}




Finally, we test our framework's performance when one or both of the similarity loss is absent.
We find that either similarity loss alone can boost the model's nighttime performance while combining them achieves the best result.

\begin{figure*}
    \centering
    \includegraphics[width=\linewidth]{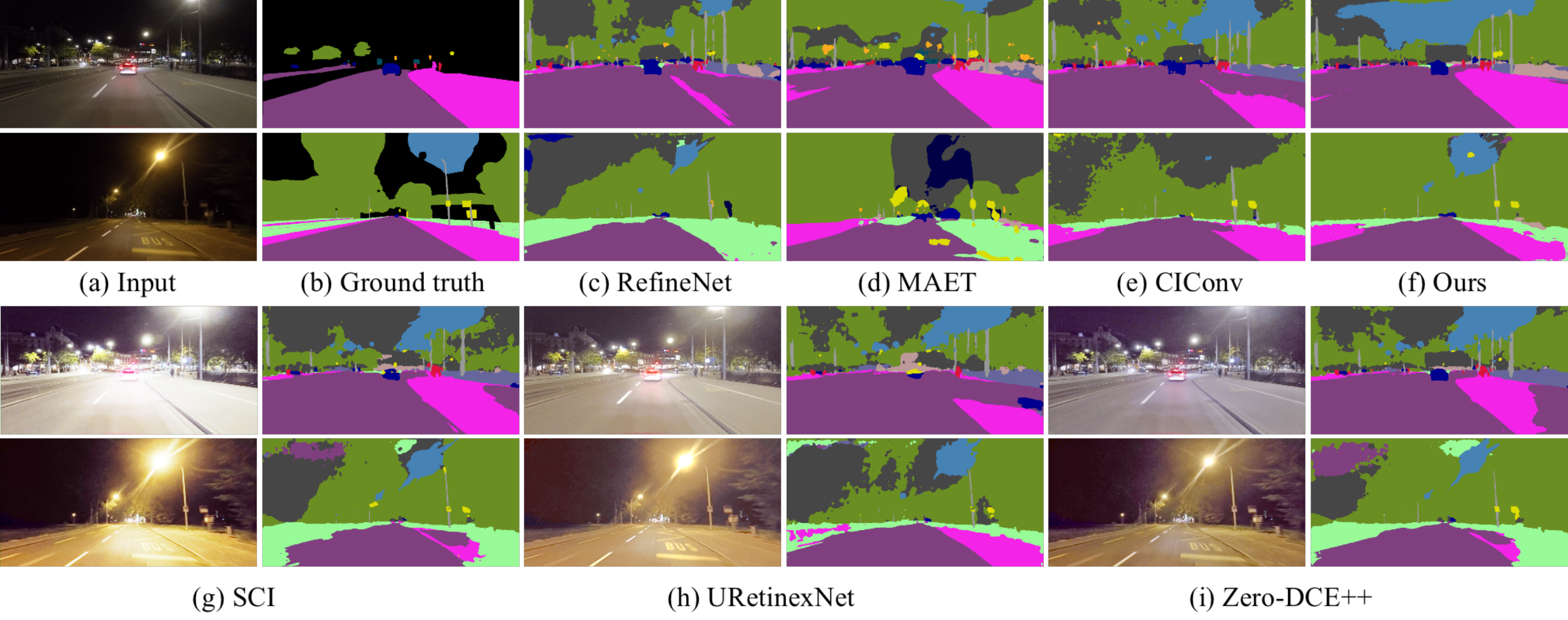}
    \caption{Semantic segmentation results. For each group, the first row: Nighttime Driving~\cite{Nighttime-Driving}, the second row: Dark-Zurich~\cite{GCMA}.}
    \label{fig:segmentation-visuals}
\end{figure*}


\subsection{Nighttime Semantic Segmentation}
\label{sec:semantic-segmentation}
Next, we explore a more challenging nighttime vision task: semantic segmentation. We adopt RefineNet~\cite{RefineNet} with ResNet-101 backbone as the baseline.
The daytime training dataset is Cityscapes~\cite{Cityscapes}, containing 2975 images for training and 500 images for validation, all with dense annotations.
The nighttime testing datasets are Nighttime Driving~\cite{Nighttime-Driving} and Dark-Zurich~\cite{GCMA}. These two datasets contain 50 coarsely annotated and 151 densely annotated nighttime street view images.

\begin{table}[t]
    \small
    \centering
    \caption{
    Semantic segmentation results on Nighttime Driving~\cite{Nighttime-Driving} and Dark-Zurich~\cite{GCMA}, reported as percentage mIoU scores.
    }
    \begin{tabular}{l|cc}
    \toprule
    Method  &\makecell[c]{Nighttime Driving} & \makecell[c]{Dark-Zurich}\\ \midrule
    RefineNet~\cite{RefineNet} & 34.3 & 30.6\\\midrule
    \multicolumn{3}{l}{\textbf{Low-Light Enhancement}}\\\midrule
    EnlightenGAN~\cite{enlightenGAN} & 25.2 & 24.9 \\
    Zero-DCE++~\cite{Zero-DCE++} & 32.7 & 28.3\\ 
    RUAS~\cite{RUAS} & 25.1 & 23.4\\
    SCI~\cite{SCI} & 28.6 & 25.7\\
    URetinexNet~\cite{URetinexNet} & 28.1 & 24.0\\ 
    LEDNet~\cite{LEDNet} & 27.6 & 26.6 \\\midrule
    \multicolumn{3}{l}{\textbf{Domain Generalization}}\\\midrule
    AdaBN~\cite{AdaBN} & 37.2 & 31.1\\
    RobustNet~\cite{RobustNet} & 33.0 & 34.5\\
    SAN-SAW~\cite{SAN-SAW} & 28.1 & 16.0 \\\midrule
    \multicolumn{2}{l}{\textbf{Zero-Shot Day-Night Domain Adaptation}} & \\\midrule
    MAET\cite{MAET} & 28.1 & 26.4\\
    CIConv~\cite{CIConv} & 41.2 & 34.5  \\
    \textbf{Ours} & \textbf{44.9} & \textbf{40.2} \\
    \bottomrule
    \end{tabular}
    \label{tab:segmentation}
\end{table}

We benchmark our method in Table~\ref{tab:segmentation}. Low-light enhancement methods yield worse results than the baseline because they perform poorly on street scenes with complex light sources. 
Domain generalization methods fail to mitigate the huge day-night domain gap, leading to unsatisfactory results. Note that RobustNet~\cite{RobustNet} adopts DeepLab-v3~\cite{Deeplab-v3} architecture, which is superior to RefineNet~\cite{RefineNet} adopted in our implementation.
%
Among zero-shot adaptation methods, MAET~\cite{MAET} injects too much noise into images, leading to severe performance degradation. CIConv yields better results, but the improvement is limited. In comparison, our approach improves the mIoU scores to 44.9\% on Nighttime Driving and 40.2\% on Dark-Zurich.

Figure \ref{fig:segmentation-visuals} shows qualitative segmentation results on two nighttime datasets. Low-light enhancement methods perform poorly on nighttime street scenes. Our method better extracts information hidden by darkness and thus generates more accurate semantic maps.

\begin{table}[t]
    \small
    \centering
    \caption{Visual place recognition results on Tokyo 24/7~\cite{Tokyo-24/7}.}
    \begin{tabular}{lc}
    \toprule
    Method & mAP (\%)\\
    \midrule
    \textbf{Zero-Shot Day-Night Domain Adaptation} & \\
    \midrule
    EdgeMAC~\cite{deep-shape-matching} & 75.9\\
    U-Net jointly~\cite{no-fear} & 79.8\\
    GeM~\cite{CNN-retrieval-human} & 85.0\\
    CIConv-GeM~\cite{CIConv} & 88.3 \\
    \textbf{Ours}-GeM  & \textbf{90.4}\\
    \midrule
    \makecell[l]{\textbf{Day-Night Domain Adaptation} \\ (night images are available for training)} & \\
    \midrule
    U-Net jointly~\cite{no-fear} & 86.5 \\
    EdgeMAC + CLAHE~\cite{no-fear} & 90.5 \\
    EdgeMAC + U-Net jointly~\cite{no-fear} & 90.0\\
    \bottomrule
    \end{tabular}
    \label{tab:retrieval}
\end{table}

\begin{figure}[t]
    \centering
    \includegraphics[width=0.95\linewidth]{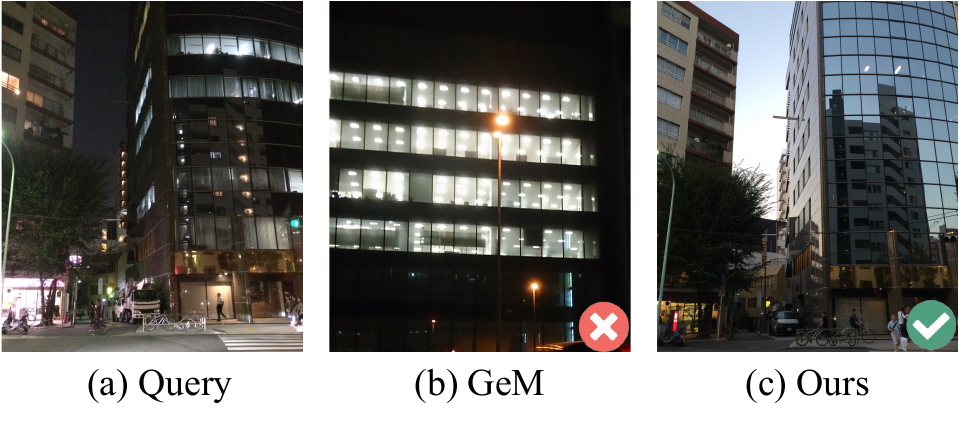}
    \caption{
    Qualitative visual place recognition results. (a) A night query from the Tokyo 24/7 dataset~\cite{Tokyo-24/7}. (b) Image retrieved by GeM~\cite{CNN-retrieval-human}. (c) Image retrieved by our method.
    }
    \label{fig:retrieval}
\end{figure}

\subsection{Visual Place Recognition at Night}
\label{sec:visual-place-recognition}

Then we explore visual place recognition, which aims to retrieve images that illustrate the same scene of a query image from an image pool. Unlike classification and segmentation, place recognition methods are not end-to-end during inference. We extend our method based on GeM~\cite{CNN-retrieval-human} with ResNet-101 backbone.
In GeM, the network receives a tuple of images $\{p,q,n_1, \cdots, n_k\}$ as input, in which the query $q$ only matches $p$.
The network is trained on a contrastive loss, similar to the model-level stage in our framework.
We retain the image-level stage and modify the model-level stage in our implementation.
We first train the darkening module $D$ as usual.
Then, we consider $D(p)$ as an additional matching for $p$, \ie, an input tuple contains two positive samples (instead of one) and $k$ negative samples.
We train our network on the Retrieval-SfM dataset~\cite{CNN-retrieval-human} and evaluate it on the Tokyo 24/7 dataset~\cite{Tokyo-24/7}, which contains city views in multiple illumination conditions and viewing directions.

Performance is reported as mean Average Precision (mAP) in Table~\ref{tab:retrieval}. Results of comparison methods are borrowed from~\cite{no-fear} and~\cite{CIConv}. Our method outperforms all zero-shot methods and is comparable to conventional domain adaptation methods.
As shown in Figure~\ref{fig:retrieval}, the baseline method gets fooled by the night's appearance, while our model finds the correct daytime image.

\subsection{Low-Light Video Action Recognition}
Although initially designed for images, our method also applies to video tasks. Here we consider an 11-class low-light video action recognition task. Normal light training data consists of 2.6k normal light video clips from HMDB51~\cite{HMDB51}, UCF101~\cite{UCF101}, Kinetics-600~\cite{Kinetics}, and Moments in Time~\cite{MiT}.
We evaluate our model on the official test split of the ARID dataset~\cite{ARID}. The action recognizer is I3D~\cite{I3D} based on 3D-ResNet~\cite{3D_ResNet}.

We extend our method to video as follows.
When training the darkening module, we input frames extracted from video clips. $\mathcal{A}$ and $\mathcal{B}$ in Eq.~(\ref{eq:darkening}) is estimated for every individual frame. We calculate $\mathcal{L}_D^{sim}$ between video clips and other losses between frames.
When generating low-light videos, frames are separately fed into the curve estimator while sharing the same exposure map $E'$.

\begin{table}[t]
    \small
    \centering
    \caption{Video action recognition results on ARID~\cite{ARID}.}
    \begin{tabular}{lc}
    \toprule
    Method & Top-1 (\%)\\
    \midrule
    I3D \cite{I3D} & 47.02 \\\midrule
    \textbf{Low-Light Video Enhancement} & \\
    \midrule
    StableLLVE \cite{StableLLVE} & 45.08 \\
    SMOID \cite{SMOID} & 47.27 \\ 
    SGZ \cite{SGZ} & 46.42\\ \midrule
    \textbf{Domain Generalization \&} &
    \\\textbf{Zero-Shot Day-Night Domain Adaptation} &\\\midrule
    AdaBN \cite{AdaBN} & 46.17\\
    \textbf{Ours} & \textbf{51.52} \\
    \bottomrule
    \end{tabular}
    \label{tab:action-recognition}
\end{table}

We report the results as Top-1 accuracy. As shown in Table~\ref{tab:action-recognition}, video enhancement methods StableLLVE~\cite{StableLLVE}, SMOID~\cite{SMOID}, and SGZ~\cite{SGZ} yield a limited performance gain. Meanwhile, our approach boosts models' performance by 4.38\%, demonstrating our superiority on videos.

\section{Conclusion}
In this paper, we propose a novel approach for zero-shot day-night domain adaptation. Going beyond a simple focus on the image-level translation or model-level adaptation, we observe a complementary relationship between two aspects and build our framework upon the similarity min-max paradigm. Our proposed method can significantly boost the model's performance at nighttime without accessing the nighttime domain. Experiments on multiple datasets demonstrate the superiority and broad applicability of our approach.

\section*{Acknowledgements}
This work is supported by the Fundamental Research Funds for the Central Universities and the National Natural Science Foundation of China under Contract No.61772043 and a research achievement of Key Laboratory of Science, Technology and Standard in Press Industry (Key Laboratory of Intelligent Press Media Technology). This research work is also partially supported by the Basic and Frontier Research Project of PCL and the Major Key Project of PCL.




{\small
\bibliographystyle{ieee_fullname}
\bibliography{egbib}
}


\end{document}